\documentclass[10pt,twocolumn,letterpaper]{article}

\usepackage{cvpr}              % To produce the CAMERA-READY version
\usepackage{bm}
\usepackage{multirow}
\usepackage{makecell}
\usepackage{graphicx}
\usepackage{amsmath}
\usepackage{amssymb}
\usepackage{booktabs}
\usepackage[table]{xcolor}
\usepackage[accsupp]{axessibility}

\usepackage[pagebackref,breaklinks,colorlinks]{hyperref}

\usepackage[capitalize]{cleveref}
\crefname{section}{Sec.}{Secs.}
\Crefname{section}{Section}{Sections}
\Crefname{table}{Table}{Tables}
\crefname{table}{Tab.}{Tabs.}

\def\cvprPaperID{4747} % *** Enter the CVPR Paper ID here
\def\confName{CVPR}
\def\confYear{2023}

\begin{document}

%%%%%%%%% TITLE - PLEASE UPDATE
% \title{Learning Better Pseudo Labels \\ For Weakly Supervised Temporal Action Localization}
\title{Improving Weakly Supervised Temporal Action Localization \\ by Bridging Train-Test Gap in Pseudo Labels}

\author{Jingqiu Zhou \textsuperscript{\rm 1}\quad
         Linjiang Huang\textsuperscript{\rm 1,2} \thanks{Corresponding author.}\quad
         Liang Wang\textsuperscript{\rm 4}
         Si Liu\textsuperscript{\rm 5}
         Hongsheng Li\textsuperscript{\rm 1,2,3}\\
\textsuperscript{\rm 1}CUHK-SenseTime Joint Laboratory, The Chinese University of Hong Kong \\
\textsuperscript{\rm 2}Centre for Perceptual and Interactive Intelligence, Hong Kong \\
\textsuperscript{\rm 3}Xidian Uniersity \\
\textsuperscript{\rm 4}Institute of Automation Chinese Academy of Science \\
\textsuperscript{\rm 5}Beihang University\\
{\tt\small 1155167063@link.cuhk.edu.hk, ljhuang524@gmail.com, hsli@ee.cuhk.edu.hk}
}

\maketitle
%%%%%%%%% ABSTRACT
\begin{abstract}
The task of weakly supervised temporal action localization targets at generating temporal boundaries for actions of interest, meanwhile the action category should also be classified. Pseudo-label-based methods, which serve as an effective solution, have been widely studied recently. However, existing methods generate pseudo labels during training and make predictions during testing under different pipelines or settings, resulting in a gap between training and testing. 
In this paper, we propose to generate high-quality pseudo labels from the predicted action boundaries. Nevertheless, we note that existing post-processing, like NMS, would lead to information loss, which is insufficient to generate high-quality action boundaries.
More importantly, transforming action boundaries into pseudo labels is quite challenging, since the predicted action instances are generally overlapped and have different confidence scores.
Besides, the generated pseudo-labels can be fluctuating and inaccurate at the early stage of training. It might repeatedly strengthen the false predictions if there is no mechanism to conduct self-correction. To tackle these issues, we come up with an effective pipeline for learning better pseudo labels. Firstly, we propose a Gaussian weighted fusion module to preserve information of action instances and obtain high-quality action boundaries. Second, we formulate the pseudo-label generation as an optimization problem under the constraints in terms of the confidence scores of action instances. Finally, we introduce the idea of $\Delta$ pseudo labels, which enables the model with the ability of self-correction. Our method achieves superior performance to existing methods on two benchmarks, THUMOS14 and ActivityNet1.3, achieving gains of 1.9\% on THUMOS14 and 3.7\% on ActivityNet1.3 in terms of average mAP. Our code is available at \url{https://github.com/zhou745/GauFuse_WSTAL.git}.
\end{abstract}

%%%%%%%%% BODY TEXT

\section{Introduction} \label{sec:intro}
The task of temporal action localization seeks to identify the action boundaries and to recognize action categories that are performed in the video. Action localization can contribute to video understanding, editing, etc. Previous works \cite{xu2017r,zhao2017temporal,lin2018bsn,chao2018rethinking,lin2019bmn} mainly solved this task in the fully supervised setting, which requires both video-level labels and frame-wise annotations. 
However, frame-wisely annotating videos is labor-intensive and time-consuming. To reduce the annotation cost, researchers start to focus on the weakly supervised setting. Considering the rich video resources from various video websites and apps, weakly supervised setting would save tremendous annotation efforts.

Unlike its supervised counterpart, the weakly supervised temporal action localization task only requires video-level category labels. The existing works mainly follow the localization-by-classification pipeline \cite{wang2017untrimmednets,zhao2020equivalent}, which trains a video-level classifier with category labels \cite{narayan2021d2}, and applies the trained classifier to each video \textbf{snippet}\footnote{We view snippets as the smallest granularity since the high-level features of consecutive frames vary smoothly over time \cite{wiskott2002slow,jayaraman2016slow}. In our work, we treat every 16 frames as a snippet}. However, due to the lack of fine-grained annotations, the model may assign high confidence to incorrect snippets such as the contextual background, which typically has a high correlation with the video-level labels, or only focus on the salient snippets, leading to incomplete localization results. There are many studies \cite{liu2019completeness,liu2021weakly,liu2021acsnet} that tried to address this discrepancy between classification and localization, and one of the promising solutions is to generate and utilize pseudo labels.

The advantage of using pseudo labels is that snippets are supervised with snippet-wise labels instead of video-level labels. Existing works \cite{pardo2021refineloc,luo2020weakly,zhai2020two,yang2021uncertainty} achieve remarkable results by introducing pseudo labels into this problem.
A commonly used strategy for generating pseudo labels is to directly utilize the temporal class activation map (TCAM) generated in previous training iterations. Nevertheless, we would like to argue that the TCAMs are not desirable pseudo labels. During testing, our goal is to obtain the action boundaries, employing the TCAMs as training targets arise the discrepancy between training and testing because they are quite different from the actual action boundaries.
An intuitive way to address this issue is to leverage the predicted action boundaries as pseudo labels. However, it is non-trivial to achieve this goal.
% First, current post-processing schemes, such as Non-Maximum Suppression (NMS), would induce much information loss of action boundaries' distribution in the final action boundaries, there are many useful action instances are filtered during post processing.
First, current post-processing schemes, such as NMS, would induce a large amount of information loss and are not sufficient to obtain high-quality action boundaries for generating effective pseudo labels.
Second, the predicted action instances are usually overlapped with each other and have different confidence scores, it is hard to assign the action categories and confidence scores for each snippet. 

To address the above issues, we propose the following two modules.
First, we propose a \textbf{Gaussian Weighted Instance Fusion} module to preserve information on the boundary distributions and produce high-quality action boundaries. Specifically, this module weightedly fuses the information of overlapped action instances. Each candidate action instance is treated as an instance sampled from a Gaussian distribution. The confidence score of each action instance is viewed as its probability of being sampled. Accordingly, we can obtain the most possible action boundaries and their confidence scores by estimating the means of Gaussians from those candidate action instances. In this way, we can produce better action boundaries, which in return help to generate more reasonable pseudo labels.

After generating high-quality action boundaries, we need to convert them into snippet-wise pseudo labels. To handle the overlapped action instances and assign snippets with proper confidence scores, we propose a \textbf{LinPro Pseudo Label Generation} module to formulate the process of pseudo-label generation as a $\ell_1$-minimization problem. First, we restrict that the average score of snippets within an action boundary should be equal to the confidence score of this action instance. This constraint guarantees that we can maintain the information of confidence scores in the generated pseudo labels. Second, snippets within an action instance might be equivalent in terms of their contribution to the confidence score. Thus we require snippet-wise scores within each action instance 
to be uniform. Based on the two constraints, we formulate the pseudo label generation as an optimization problem and solve it to obtain pseudo labels that are consistent with our predicted action boundaries. 

Furthermore, there is still one problem regarding to the use of pseudo labels. Since the generated pseudo labels can be fluctuating and inaccurate at early stage of training, without a proper self-correction mechanism, the model would keep generating wrong pseudo labels of high confidences at later training stages. To address this issue, we propose to utilize the \textbf{$\Delta$ pseudo labels}, instead of the original pseudo labels, as our training targets. We 
calculate the difference between the pseudo labels of consecutive training epochs as the $\Delta$ pseudo labels. 
In general, the model would provide more accurate predictions along with the training.
In this way, the model will update its predictions toward the class with the confidence increasing instead of the class with the largest pseudo label value, and thus empowers the model with the ability of self-correction.

% What's more, there is still one problem regarding to the use of pseudo label. \textcolor{red}{When directly used as the training target, pseudo labels may self-reinforce some early-stage false positive predictions \cite{kim2021self}. Without a proper self-correction mechanism, the model performance would be hindered by such pseudo labels. To address this issue, we propose to use the \textbf{$\Delta$ pseudo labels}, instead of using the generated pseudo label as our training target. We compute the changes of the pseudo label between consecutive training epochs. In this way, the model will move its logits toward the class with the most positive change instead of the class with largest pseudo-label value. For this reason, our $\Delta$ pseudo label can correct some overconfident false positive pseudo labels.}

% In this way, if the model becomes less confident about a prediction, it will result in a negative value in the $\Delta$ pseudo label. In return, this $\Delta$ pseudo label will decrease the logits score of . For this reason, our $\Delta$ pseudo label can correct some mistakes of pseudo label made at early training stage.

The contribution of this paper is four-fold.
(a) We propose a Gaussian Weighted Instance Fusion module, which can effectively generate high-quality action boundaries.
(b) We propose a novel LinPro Pseudo Label Generation strategy by transforming the process of pseudo-label generation into a $\ell_1$-minimization problem.
(c) We propose to utilize $\Delta$ pseudo labels to enable model with self-correction ability for the generated pseudo labels. 
(d) Compared with state-of-the-art methods, the proposed framework yields significant improvements of \textbf{1.9\%} and \textbf{3.7\%} in terms of average mAP on THUMOS14 and ActivityNet1.3, respectively.

\section{Related Work} \label{sec:related_work}
For the task of weakly supervised temporal action localization, the discrepancy between classification and localization has been observed by many researchers \cite{liu2019completeness,moniruzzaman2020action,narayan2021d2}. To alleviate this problem, many efforts have been made. We divide these methods into the following four categories: metric learning-based methods, erasing-based methods, multi-branch methods, and pseudo-label-based methods. 

For the metric learning-based methods, previous works like W-TALC \cite{paul2018w}, 3C-Net \cite{narayan20193c}, RPN \cite{huang2020relational} and A2CL-PT \cite{min2020adversarial} utilize the center loss \cite{wen2016discriminative}, clustering loss \cite{yang2018robust} and triplet loss \cite{schroff2015facenet}. In general, these methods \cite{paul2018w,huang2020relational} tend to obtain video-level features from the most discriminative snippets features. 
Therefore, these methods failed to capture snippets of less-discriminative features. Although several methods like RSKP \cite{rskp} have tried to address this issue by propagating the knowledge of representative snippets, this method is also a double-edge sward. The knowledge propagation is a bi-lateral process, while we are passing information from discriminative snippets to less-discriminative ones, the transverse also takes place, which may hinder the model from learning high confidence snippets.

Another category is the erasing-based methods. These methods \cite{zhong2018step,min2020adversarial} embrace the idea of exploring less-discriminative features from erasing discriminative ones. Originated from the adversarial complementary learning \cite{zhang2018adversarial}, these methods repeatedly find the most discriminative snippets and erase them. However, it is difficult to set proper thresholds for different classes with different complexities.

The third category mainly adopts the multi-branch architecture. These methods \cite{liu2019completeness,liu2021weakly,liu2021acsnet,islam2021hybrid,huang2021foreground} share the similar idea as the erasing-based methods, while they differ from those methods by parallel processing. Likewise, it shares the same problem with the erasing-based methods.

The last category is the pseudo-label-based methods. These methods originated from Refine-Loc \cite{pardo2021refineloc}, which generates snippet-level hard pseudo labels. Then various works follow this idea and try to generate more accurate pseudo labels. For example, works like EM-MIL \cite{luo2020weakly}, RSKP \cite{rskp} fit the pseudo-label generation into an expectation-maximization framework. UGCT \cite{yang2021uncertainty} uses an uncertainty-guided collaborative training strategy. However, most of these methods utilize the TCAM or variants of TCAM as the pseudo label, leading to discrepancy between training and testing. Besides, these methods cannot address the problem that pseudo labels can be inaccurate at the early training stage. And the inaccurate pseudo labels will keep generating wrong pseudo labels without the self-correction mechanism. In contrast to existing pseudo-label-based methods, we propose to generate pseudo labels from action boundaries, alleviating the discrepancy between training and testing. Besides, the introduction of $\Delta$ pseudo labels enables the model with the ability of self-correction.

\section{Method} \label{sec:proposed_method}
\begin{figure*}[t]
\centering
\includegraphics[width=0.88\textwidth]{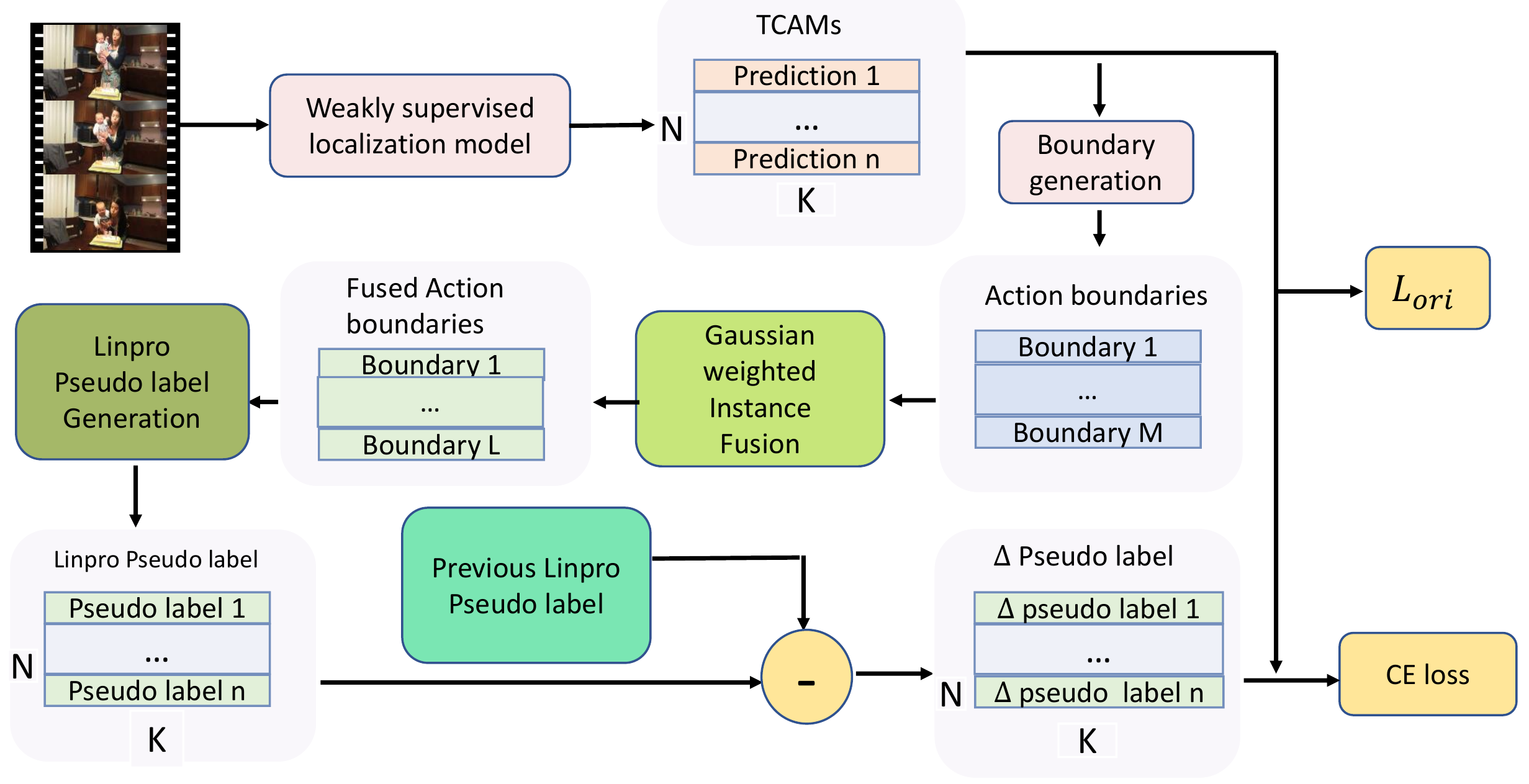}
\caption{The overview of our method. We first feed the video into a weakly supervised action localization model to generate the temporal class activation map (TCAM). The obtained TCAM is transformed into action boundaries using multiple thresholds. We utilize the Gaussian Weighted Instance Fusion module to merge the candidate action instances into high-quality action boundaries. Then, we transform the action boundaries into snippet-wise pseudo labels by solving a linear programming problem. To enable the model with the ability of self-correction, we utilize the $\Delta$ pseudo labels, which are the differences between two pseudo labels generated at consecutive epochs, as the final pseudo labels. Note that, all the original losses of the weakly supervised localization model are preserved.}
\label{fig:framework}
\vspace{-3mm}
\end{figure*}

In this section, we detail the proposed method. An overall illustration of the pipeline is demonstrated in Figure \ref{fig:framework}.

\vspace{-4mm}
\paragraph{Problem definition.}
% Suppose \(V=\{v_i\}_{i=1}^{l}\) be a video of temporal length \(l\). Each video is divided into a series of non-overlapping snippets. Assume that we have a set of \(N\) training videos \(\{V_i\}_{i=1}^{N}\) being annotated with their action categories \(\{\bm{y}_i\}_{i=1}^{N}\), where \(\bm{y}_i\) is a binary vector indicating the presence/absence of each of $K$ action classes. During inference, for a video, our target is to predict a set of action instances \(\{(c, q, s, e)\}\), where \(c\) denotes the predicted action class, \(q\) is the confidence score, \(s\) and \(e\) represent the start time and end time of the action instance.

Given \(N\) training videos \(\{V_i\}_{i=1}^{N}\), during training, we can only access their action categories \(\{\bm{y}_i\}_{i=1}^{N}\), where \(\bm{y}_i\) is a binary vector indicating the presence/absence of each of $K$ action classes. During inference, for a video, our target is to predict a set of action instances \(\{(c, q, s, e)\}\), where \(c\) denotes the predicted action class, \(q\) is the confidence score, \(s\) and \(e\) represent the start time and end time of the action instance.

\vspace{-5mm}
\paragraph{Overview.}
Our target is to narrow down the gap between training and testing by generating pseudo labels from the predicted action boundaries. First, we propose a Gaussian Weighted Instance Fusion module to obtain high-quality action boundaries. 
This module serves as an effective alternative to the non-maximum suppression (NMS), which weightedly fuses action instances that are overlapped with each other, rather than filtering low-confident instances.
% Then a post-processing is applied to estimate the hyper-parameters of a template distribution from those potential action instances. The outputting post processed action tuples fuse information from multiple overlapping potential action instances.

% Then the more informative action instances are used to generate pseudo-labels. This novel pseudo-label generation paradigm is important in narrow down the gap between training and testing. A large number of previous works have utilized pseudo labels for compensating the weakness of video level annotation in snippet level predictions, while they normally ignore the discrepancy between their generated training pseudo-labels and testing action instances.

After generating the high-quality action instances, we need to transfer them into snippet-wise pseudo labels. The key idea is generating pseudo labels that can preserve the confidence score within each action instance and make the scores within an action instance as uniform as possible. Therefore, we formulate it as a linear programming problem with the above two constraints to obtain the optimal snippet-wise pseudo labels.

Lastly, we propose the idea of $\Delta$ pseudo labels for better utilization of the generated pseudo labels. Instead of naively using the generated pseudo labels as training target, we take the differences of pseudo labels between two consecutive epochs as the final pseudo labels. In this way, we can capture the confidence change of the model for the pseudo labels, which represents the pseudo labels' reliability, and thus enable the model with the ability of self-correction.

\vspace{-5mm}
\paragraph{Feature extraction and network design.}
Following previous works \cite{nguyen2018weakly,paul2018w}, for each video $V_i$, every $16$ consecutive frames is processed into a snippet-level feature. In our case, the Inflated 3D (I3D) \cite{carreira2017quo} pre-trained on the Kinetics-400 dataset \cite{kay2017kinetics} is used to encode the video snippets. The output snippet-level features are in $\mathbb{R}^{2048}$, thus we convert the video of $l$ snippets into a feature matrix of $\mathcal{F} \in \mathbb{R}^{l \times 2048}$. 
As a plug-in method, our method can be applied to most of the existing approaches. Here, we adopt RSKP \cite{rskp} as the backbone to obtain the temporal class activation map.

% \textcolor{red}{The network architecture is adapted from \cite{rskp}. In previous works, the FAC-Net \cite{huang2021foreground} is choosing as the classification head which generate a temporal class activation map(TCAM). Then a multi-threasholding strategy is applied to transform TCAM into action instances of interest.}

In the sections below, we first describe how to fuse candidate action instances to obtain high-quality action boundaries, then we elaborate on the generation of pseudo labels, and finally, we introduce our $\Delta$ pseudo labels.
% \textcolor{blue}{With a little abuse of notation, we omit the subscript $i$ for video number in the following sections.}

\subsection{Gaussian Weighted Instance Fusion} \label{sec:Prediction Fusion}
The classification head predicts labels for each individual snippet, it results in temporal class activate map (TCAM) $L \in \mathbb{R}^{l \times K}$, where $l$ is the length of a given video and $K$ is the number of classes. There are multiple ways of transforming the TCAM into action instances of interest. A commonly used way is employing multiple thresholds \cite{lee2020background,huang2021foreground,huang2021modeling} to obtain redundant action instances and then resort to the non-maximum suppression (NMS) for duplicating highly-overlapped instances. 

% The generated action instances are all intervals along the temporal dimension within which all the snippets score is greater than the threshold. \textcolor{blue}{This method is proposed by Lee\cite{lee2020background}, and we followed their work and generated the action prediction set $P_i$ for each video $V_i$.}

% Now we intend to further refine these predictions. Because each class is treated by exactly the same procedure, thus we only describe class $c$. 

 Traditional NMS is designed to only pick out the predicted action instances with high confidence scores, while throwing away those low-confident action instances. However, due to the weakly supervised setting, the confidence score lacks explicit supervision by snippet-level annotations. It is unwise to just trust the predictions with the high confidence scores and suppress other predictions with low confidence scores. Moreover, those suppressed action instances indicates different level of confidence at different locations. Thus large amount of information is lost due to the NMS procedure. To address the above issues, we present a novel Gaussian weighted fusion module to aggregate all the action instances who were suppressed by NMS.

% $p_{i,l}=(c,q_l,t_{s,l},t_{e,l})$ from the Prediction set $P_i$. In our weakly supervised setting, the confidence score lacks supervision directly from snippets level annotations. It is unwise to just trust the prediction with the highest confidence scores and suppress other predictions with lower score confidence. Then we present the following prediction fusion module. 

Suppose for a class $c$, we have $M$ candidate action instances $\mathbf{A}=\{a_1,\cdots,a_M\}$. Each action instance is of the $a_{i}=(c_i,q_i,s_i,e_i)$ where $c_i$ is the predicted class, $q_i$ is the confidence score, and $s_i,e_i$ are the boundaries.  We collect all the prediction instances, whose IoUs are greater than a predefined threshold $h_{fuse}$ to the most confident predicted instance $a_{*}$ in set $\mathbf{A}$. The index set of these prediction segments are denoted as $\mathcal{I}_{*}$
\begin{equation}
    \mathcal{I}_{*}=\{k|\text{IoU}(a_{k},a_{*})>h_{fuse}\}.
\end{equation}
Assuming the confidence scores $\{q_i\}$, start points $\{s_i\}$ and end points $\{e_i\}$ of the collected instances in index set $\mathcal{I}_{*}$ satisfy independent Gaussian distribution, whose probability density function (PDF) can be formulated as:
\begin{equation}
    N({\blacktriangle})=\frac{1}{\sigma_{{\blacktriangle}}\sqrt{2\pi}} \exp(-\frac{({\blacktriangle}-\mu_{\blacktriangle})^2}{2\sigma_{\blacktriangle}^2}),
\end{equation}
where ${\blacktriangle} \in \{q,s,e\}$. Now we tend to match the template Gaussian distribution as close as to our sampled predictions. Note that each action instance has a confidence score, if we regard the confidence scores as the un-normalized logits of action instances, then we can compute the probability of sampling the $k$-th action instance as:
\begin{equation}
    g_k = \frac{\exp{(q_k/T)}}{\sum_{i\in \mathcal{I}_{*}}\exp{(q_i/T)}},
\end{equation}
where $T$ is a temperature hyper-parameter. By minimizing the cross-entropy between $N(\blacktriangle)$ and $\{g_k\}$ we have:
\begin{equation}
    \mu_{\blacktriangle} = \sum_{i \in \mathcal{I}_{*}}{\blacktriangle}_i g_i,
\end{equation}
where $\blacktriangle$ stands for $\{q,s,e\}$. Therefore, we can obtain the weightedly fused value $\mu_{q}$, $\mu_{s}$ and $\mu_{e}$, which are taken as the confidence score, start point and end point for the fused action instance. Then we delete all action instances of interest from $\mathcal{I}_{*}$ in $\mathbf{A}$, and repeat the above process until $\mathbf{A}$ is empty.
Here we also tried some other distribution templates, such as exponential distribution, the Gaussian distribution achieves the best performance among them. 

\subsection{LinPro Pseudo Label Generation}

After generating the high-quality action boundaries, we propose to transform them into snippet-wise pseudo labels. In this way, the video snippets are directly supervised by the final targets, \ie, action boundaries, instead of the TCAM, and thus close up the gap between testing and training.
A naive approach to achieve this is to directly assign a hard label to each snippet within the post-processed action instance.
However, this naive approach suffers from several drawbacks. First, snippets from different action instances may have different confidence scores, assigning them with the same hard label brutally ignores this difference. 
Second, using hard labels actually overlooks the importance of the predicted confidence scores, which could represent the instances' quality measured by the model to some extent.
Third, even after post-processing, some predicted instances are still overlapped, this naive approach will simply merge them together and result in false predictions. 

To address the above issues, we propose an optimization-based method to generate desirable pseudo-labels. Since the confidence score are important information action instances carry on, we intend to find the pseudo label which can characterize the property of the confidence score of each action instance. To this end, we formulate this optimization problem based on two constraints: (1) the generated pseudo label should preserve the confidence score of each action instance. (2) The scores of the generated pseudo labels should be uniform. The first constraint is easy to comprehend because we intend to maintain the confidence score of each action instance. As for the second constraint, it follows from the fact that most of the snippets within an action instance should be equivalent for classification and localization.

% Following Inner-Outer Contrast score \cite{shou2018autoloc,lee2020background}, the difference between the inner boundary's average score and the inflated outer boundary's average score is viewed as the confidence score $q$ of each action instance, which can be formulated as:
% \begin{equation}
%     q = \rm{score}_{\rm{inner}} - \rm{score}_{\rm{outer}}
% \end{equation}

% For a predicted action instance of category $c$, its inner score and outer score can be computed as :
% \begin{equation}
% \begin{aligned}
%     \rm{score}_{\rm{inner}}&=\frac{\sum_{i=s}^{e} L_{i,c}}{e-s} , \\
%     \rm{score}_{\rm{outer}}&=\frac{\sum_{i=s-\alpha (e-s)}^{s}L_{i,c} + \sum_{i=e}^{e+\alpha (e-s)}L_{i,c}}{\alpha (e-s)},
% \end{aligned}
% \end{equation}
% where $\alpha$ is hyper-parameter controlling the inflated proportion of the outer boundaries, $s,e$ are the starting and ending points of an action instance,  and $L\in \mathbb{R}^{l\times K}$ is the TCAM of a length $l$ video. 

\begin{figure}[!t]
  \centering
  \includegraphics[width=0.9\columnwidth]{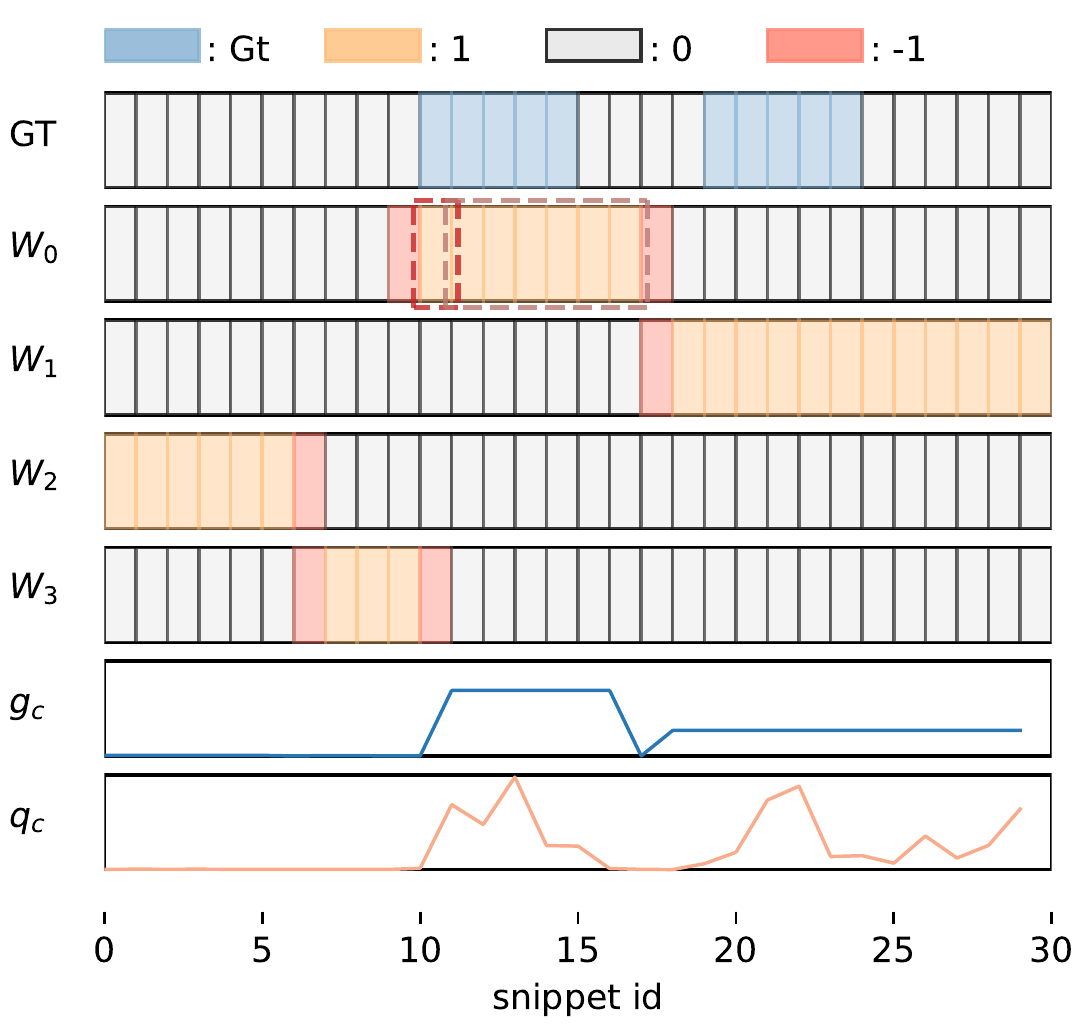}
  \caption{Illustration of the process of pseudo label generation. GT is the ground truth snippet-wise label, where the blue snippet denotes the action locations. $W_0, W_1, W_2, W_3$ are four constraints generated from four action instances. The yellow, white, and red snippets denote the constraints weights of $1, 0, -1$ respectively. $q_c$ is the predicted snippet-wise score, and $g_c$ is the generated pseudo label. The snippets in each dashed rectangle are equivalent.} \label{fig:pseudo_label}
\vspace{-4mm}
\end{figure}

To address the first constraint, we convert each action instance into a linear constraint. 
Specifically, given the $n$ action instances of the class $c$, we first construct $n$ weighting vectors $\{W_j\in \mathbb{R}^l | j=1 \dots n\}$, where $l$ is the length of the video, each of which serves as the constraint for the corresponding action instance. Here, we follow Inner-Outer Contrast score \cite{shou2018autoloc,lee2020background} to construct the weighting vectors, by dividing the whole video into the inner areas, outer areas, and background. In specific, for $W_j$, we set its $i$-th element $W_{j,i}$ as $1$ if the $i$-th snippet is in the inner area of action instance $j$, \ie, $s_j\leq i\leq e_j$. Likewise, we set it as $-1$ if it lies in the outer area, \ie, $s_j-\alpha  (e_j-s_j) \leq i<s_j$ or $e_j \leq i<e_j+\alpha (e_j-s_j))$, otherwise 0 for background. This process is detailedly shown in Figure \ref{fig:pseudo_label}. From such a construction, we have the following linear equations for the $n$ action instances:
\begin{equation}
\begin{gathered}
    q_{c,1} = W_{1}^{T}g_{c}, \\
    \vdots \\
    q_{c,n} = W_{n}^{T}g_{c}, \\
\end{gathered}
\end{equation}
where $g_c\in \mathbb{R}^l$ is the pseudo label to be generated for the class $c$, and $q_{c,i}$ is the confidence score of class $c$ for the $i$-th action instance. To this end, we constrain the generated pseudo label to maintain the Inner-Outer Contrast scores of action instances. Besides, the larger the value of $\sum_{j=1}^{n} W_{j,i}$, the more important the $i$-th snippet is. It represents the $i$-th snippet lies in more inner areas and fewer outer areas. Intuitively, we want to generate the pseudo label, which has a larger value at more important snippets. To achieve this, we minimize the $\ell_1$-norm of the pseudo-label $g_c$. The optimization problem can be finally formulated as:
\begin{equation}
    \begin{aligned}\label{eq1}
    \hat{g}_{c} &= \textbf{argmin}{\|g_{c}\|_1}\\ 
    \textbf{s.t.}\;q_{c} &=  W^{T}g_{c} \\
    g_c&\geq 0,
\end{aligned}
\end{equation}
where $q_c=(q_{c,1},\dots,q_{c,n})$ and $W=[W_1; \dots ;W_n] \in \mathbb{R}^{l \times n}$.
Even though, the above optimization target can achieve minimizing $\ell_1$-norm of $g_c$, it does not guarantee a uniform-shaped pseudo label. To satisfy the second constraint, we first collect the snippets, which are equivalent for the optimization problem. Here, we define the two snippets $i$, $j$ as equivalent, if the $i$-th and the $j$-th rows of $W$ are the same, which means they fall on the same inner area/outer area/background area, or the same overlapped area. For those equivalent snippets, we average their scores in the solved pseudo label $\hat{g}_{c}$ of Eq. (\ref{eq1}), and re-assign them with the average value. In this way, we transform the pseudo label $\hat{g}_{c}$ into a more uniform form while satisfying the optimization target of Eq. (\ref{eq1}).

For each of the $K$ classes, we conduct the above procedure to obtain its pseudo label. After that, we concatenate all the pseudo labels of $K$ classes together into the final pseudo label $G\in \mathbb{R}^{l\times K}$ for the input video. 

% Note that solving the optimization problem requires some time, thus we do not update our pseudo label at every iteration. In stead, we choose to update the pseudo label at some given checkpoints.

\subsection{$\Delta$ Pseudo Label}
Although the generated pseudo labels can capture the boundaries of predicted action instances and preserve the information of confidence scores, they still have certain drawbacks when used directly. If we apply our pseudo label $G$ directly to the TCAM $L$ with cross-entropy (CE) loss, \ie, $ \mathcal{L}_{ce} = \sum \text{CE}(\text{softmax}(L_j),G_j)$, where $j$ denotes the $j$-th snippet, the negative gradient of the CE loss with respect to the snippet $j$ and the class $c$ is:
% \begin{equation}
%     -grad_{j} = s*(\hat{G}_{j}-P_{j}),
% \end{equation}
\begin{equation}
    - \frac{\partial \mathcal{L}_{ce}}{\partial P_{j,c}} = G_{j,c}(1-P_{j,c}),
\end{equation}
where $P_{j}=\text{softmax}(L_{j})$. Since $(1-P_{j,c})$ is always positive, this negative gradient indicates that we are pushing the TCAM towards the direction of $G_{j,c}$, which is also always positive. What's worse, at the early stage of training, the generated pseudo labels are usually fluctuating and inaccurate, it would force the model to keep generating wrong pseudo labels of high confidence at later training stages.

To alleviate this issue, we propose to utilize the $\Delta$ pseudo labels. In specific, we take the difference of two pseudo labels, which are collected at two consecutive epochs of pseudo label generation, as the $\Delta$ pseudo label, which can be obtained as:
\begin{equation}
    \Delta G^t=G^{t}-G^{t-1},
\end{equation}
where upper-script $t$ stands for the pseudo label generated at the $t$-th time. Now, applying the cross entropy loss, the negative gradient turns into:
\begin{equation}
    - \frac{\partial \mathcal{L}_{ce}}{\partial P_{j,c}} = \Delta G_{j,c}^t(1-P_{j,c}).
\end{equation}
% \begin{equation}
%     -grad_{j} = s*(\Delta \hat{G}^{t}_{j}-P_{j}).
% \end{equation}
In this way, the training loss is driving the TCAM to the direction of $\Delta {G}_{j,c}^t$, which could be either positive or negative. 
Since the model in general tends to provide more accurate predictions when the training goes on, it can be expected that when the snippet $j$ is a part of the action category $i$, its $\Delta {G}^t_{j,c}$ could be positive. In contrast, if the snippet $j$ does not belong to the class $i$, its $\Delta {G}^t_{j,c}$ could be negative since the model is prone to lower its score. Therefore, this mechanism mitigates the issue of generating inaccurate pseudo labels at the early stage of training, by considering the relative changes of pseudo labels rather than the absolute values, and thus empowers the model with the ability of self-correction, so as to obtain better performance.

\section{Experiments} \label{sec:experiments}
\subsection{Datasets}

\vspace{-1mm}
\paragraph{THUMOS14 \cite{THUMOS14}.}
THUMOS14 is a dataset with 20 classes of actions. In our experiments, we consider of subset of THUMOS14 which has frame-wise annotations for all 20 classes of actions. We train our model on the 200 validation video and evaluate it on the 212 testing video. Note that we do not use the frame-wise label at training stage.

\vspace{-5mm}
\paragraph{ActivityNet1.3 \cite{caba2015activitynet}.}
ActivityNet1.3 is a dataset that contains 200 daily activities. This dataset provides 10,024 videos for training, 4,926 validation videos and 5,044 testing ones. Our model is trained on the training set and tested on the validation set.

\subsection{Implementation Details} \label{sec:implementation_datails}
\vspace{-1mm}
\paragraph{Model details.}
We pre-process each video into snippets, which are extracted into 2048-$d$ features by the I3D model pre-trained on Kinetics-400\cite{carreira2017quo}. Then we strictly follow the network design of \cite{rskp} for model designing. 

\vspace{-5mm}
\paragraph{Training details.}
Our method is trained with a mini-batch size of 10 and 128 with Adam \cite{kingma2014adam} optimizer for THUMOS14 and ActivityNet1.3, respectively. The hyper-parameter of temperature $T$ for Gaussian weighted fusion is set as $0.1$. At the early training stage, the model is insufficient to generate high quality pseudo labels, thus we start to generate pseudo labels from epoch 200, and renew the pseudo label at epochs 215, 230, 245, 270 and 290. The training procedure stops at 350 epochs with the learning rate $5 \times 10^{-5}$. Unless state otherwise, this is our default training setting in the following experiments.

\vspace{-5mm}
\paragraph{Testing details.}
The whole sequence of a video is used as testing input. Our model produces snippet-level predictions, and we simply up-sample the predictions to match the original frame rate.
Following \cite{lee2020background}, we use a set of thresholds to obtain the predicted action instances. After that, we apply our Gaussian weighted fusion strategy instead of non-maximum suppression to generate the more accurate action instances. During testing, the hyper-parameter of temperature $T$ for Gaussian weighted fusion is set as $0.03$.
\begin{table}[!t]
\begin{center}
\caption{Results on ActivityNet1.3 validation set. AVG indicates the average mAP at IoU thresholds 0.5:0.05:0.95.}
\vspace{-3mm}
\label{table:activity1.3}
\begin{tabular}{l|cccc}
\hline
\hline
\multirow{2}{*}{Method} & \multicolumn{4}{c}{mAP @ IoU} \\
\cline{2-5}
& 0.5 & 0.75 & 0.95 & AVG \\
\hline
\hline
%R-C3D \cite{xu2017r} & 26.8 & - & - & 12.7 \\
TAL-Net  \cite{chao2018rethinking} & 38.2 & 18.3 & 1.3 & 20.2\\
BSN \cite{lin2018bsn} & 46.5 & 30.0 & 8.0 & 30.0\\
GTAN \cite{long2019gaussian} & 52.6 & 34.1 & 8.9 & 34.3\\
\hline
%STPN (I3D) \cite{nguyen2018weakly} & 29.3 & 16.9 & 2.6 & -\\
%CMCS (I3D) \cite{liu2019completeness} & 34.0 & 20.9 & 5.7 & 21.2\\
%BM (I3D) \cite{nguyen2019weakly} & 36.4 & 19.2 & 2.9 & -\\
BaS-Net (I3D) \cite{lee2020background} & 34.5 & 22.5 & 4.9 & 22.2\\
A2CL-PT (I3D) \cite{min2020adversarial} & 36.8 & 22.0 & 5.2 & 22.5\\
ACM-BANet (I3D) \cite{moniruzzaman2020action} & 37.6 & 24.7 & 6.5 & 24.4\\
TSCN (I3D) \cite{zhai2020two} & 35.3 & 21.4 & 5.3 & 21.7\\
WUM (I3D) \cite{lee2021weakly} & 37.0 & 23.9 & 5.7 & 23.7\\
%LES (I3D) \cite{liu2021weakly} & 35.1 & 23.7 & 5.6 & 23.2 \\
TS-PCA (I3D) \cite{liu2021blessings} & 37.4 & 23.5 & 5.9 & 23.7 \\
UGCT (I3D) \cite{yang2021uncertainty} & 39.1 & 22.4 & 5.8 & 23.8 \\
AUMN (I3D) \cite{luo2021action} & 38.3 & 23.5 & 5.2 & 23.5\\
FAC-Net (I3D) \cite{huang2021foreground} & 37.6 & 24.2 & 6.0 & 24.0 \\
RSKP (I3D)\cite{rskp} & 40.6 & 24.6 & 5.9 & 25.0 \\
ASM-Loc (I3D)\cite{he2022asm} & 41.0 & 24.9 & 6.2 & 25.1 \\
\textbf{Ours} &\textbf{43.4} & \textbf{28.8} & \textbf{9.9} &\textbf{28.8}\\
\hline
\hline
\end{tabular}
\vspace{-5mm}
\end{center}
\end{table}

\begin{table*}[!t]
\begin{center}
\caption{Comparisons of detection performance on THUMOS14. UNT and I3D represent UntrimmedNet features and I3D features, respectively. $\dag$ means that the method utilizes additional weak supervisions, \eg, action frequency.}
\label{table:THUMOS14}
\vspace{-3mm}
\resizebox{2\columnwidth}{!}{
\begin{tabular}{c|l|c|ccccccc|ccc}
\hline
\hline
\multirow{2}{*}{Supervision} & \multirow{2}{*}{Method} & \multirow{2}{*}{Feature} & \multicolumn{7}{c|}{mAP @ IoU (\%)} & \multirow{2}{*}{\makecell{AVG\\(0.1:0.5)}} & \multirow{2}{*}{\makecell{AVG\\(0.3:0.7)}} & \multirow{2}{*}{\makecell{AVG\\(0.1:0.7)}} \\
\cline{4-10}
&  &  & 0.1 & 0.2 & 0.3 & 0.4 & 0.5 & 0.6 & 0.7 &  &  & \\
\hline
\hline
\multirow{3}{*}{Full}
%& S-CNN \cite{shou2016temporal}, CVPR2016 & - & 47.7 & 43.5 & 36.4 & 28.7 & 19.0 & 10.3 & 5.3 & 35.0 & 19.9 & 27.3 \\
%& R-C3D \cite{xu2017r}, ICCV2017 & - & 54.5 & 51.5 & 44.8 & 35.6 & 28.9 & - & - & 43.1 & - & - \\
& SSN \cite{zhao2017temporal} (ICCV'17) & - & 60.3 & 56.2 & 50.6 & 40.8 & 29.1 & - & - & 49.6 & - & -\\
%& 2018 & TAL-Net \cite{chao2018rethinking} & 59.8 & 57.1 & 53.2 & 48.5 & 42.8 & 33.8 & 20.8 & 45.1 \\
& BSN \cite{lin2018bsn}  (ECCV'18) & - & - & - & 53.5 & 45.0 & 36.9 & 28.4 & 20.0 & - & 36.8 & - \\
%& BMN \cite{lin2019bmn}, ICCV2019 & - & - & - & 56.0 & 47.4 & 38.8 & 29.7 & 20.5 & - & 38.5 & - \\
& GTAN \cite{long2019gaussian} (CVPR'19) & - & 69.1 & 63.7 & 57.8 & 47.2 & 38.8 & - & - & 55.3 & - & -\\
\hline
\hline
\multirow{2}{*}{Weak $\dag$}
& STAR \cite{xu2019segregated}  (AAAI'19) & I3D & 68.8 & 60.0 & 48.7 & 34.7 & 23.0 & - & - & 47.0 & - & -\\
& 3C-Net \cite{narayan20193c}  (ICCV'19) & I3D & 59.1 & 53.5 & 44.2 & 34.1 & 26.6 & - & 8.1 & 43.5 & - & -\\
\hline
\hline
\multirow{20}{*}{Weak}
%& UntrimmedNet \cite{wang2017untrimmednets}, CVPR2017 & - & 44.4 & 37.7 & 28.2 & 21.1 & 13.7 & - & - & 29.0 & - & - \\
%& Hide-and-Seek \cite{singh2017hide}, ICCV2017 & - & 36.4 & 27.8 & 19.5 & 12.7 & 6.8 & - & - & 20.6 & - & - \\
%& Zhong \emph{et al}. \cite{zhong2018step}, MM2018 & - & 45.8 & 39.0 & 31.1 & 22.5 & 15.9 & - & - & 30.9 & - & - \\
%& AutoLoc \cite{shou2018autoloc}, ECCV2018 & UNT & - & - & 35.8 & 29.0 & 21.2 & 13.4 & 5.8 & - & 21.0 & - \\
& CleanNet \cite{liu2019weakly} (ICCV'19) & UNT & - & - & 37.0 & 30.9 & 23.9 & 13.9 & 7.1 & - & 22.6 & - \\
%& STPN \cite{nguyen2018weakly}, CVPR2018 & I3D & 52.0 & 44.7 & 35.5 & 25.8 & 16.9 & 9.9 & 4.3 & 35.0 & 18.5 & 27.0 \\
%& W-TALC \cite{paul2018w}, ECCV2018 & I3D & 55.2 & 49.6 & 40.1 & 31.1 & 22.8 & - & 7.6 & 39.8 & - & - \\
%& MAAN \cite{yuan2019marginalized}, ICLR2019 & I3D & 59.8 & 50.8 & 41.1 & 30.6 & 20.3 & 12.0 & 6.9 & 40.5 & 22.2 & 31.6 \\
%& CMCS \cite{liu2019completeness}, CVPR2019 & I3D & 57.4 & 50.8 & 41.2 & 32.1 & 23.1 & 15.0 & 7.0 & 40.9 & 23.7 & 32.4 \\
%& BM \cite{nguyen2019weakly}, ICCV2019 & I3D & 60.4 & 56.0 & 46.6 & 37.5 & 26.8 & 17.6 & 9.0 & 36.3 \\
%& BaS-Net \cite{lee2020background}, AAAI2020 & I3D & 58.2 & 52.3 & 44.6 & 36.0 & 27.0 & 18.6 & 10.4 & 43.6 & 27.3 & 35.3 \\
% & RPN \cite{huang2020relational} (AAAI'20) & I3D & 62.3 & 57.0 & 48.2 & 37.2 & 27.9 & 16.7 & 8.1 & 46.5 & 27.6 & 36.8 \\
%& DGAM  \cite{shi2020weakly}, CVPR2020 & I3D & 60.0 & 54.2 & 46.8 & 38.2 & 28.8 & 19.8 & 11.4 & 37.0 \\
& TSCN \cite{zhai2020two} (ECCV'20) & I3D & 63.4 & 57.6 & 47.8 & 37.7 & 28.7 & 19.4 & 10.2 & 47.0 & 28.8 & 37.8 \\
& EM-MIL \cite{luo2020weakly} (ECCV'20) & I3D & 59.1 & 52.7 & 45.5 & 36.8 & 30.5 & 22.7 & \textbf{16.4} & 45.0 & 30.4 & 37.7 \\
& A2CL-PT \cite{min2020adversarial} (ECCV'20) & I3D & 61.2 & 56.1 & 48.1 & 39.0 & 30.1 & 19.2 & 10.6 & 46.9 & 29.4 & 37.8 \\
%%& ACM-BANet \cite{moniruzzaman2020action}, MM2020 & I3D & 64.6 & 57.7 & 48.9 & 40.9 & 32.3 & 21.9 & 13.5 & 48.9 & 31.5 & 39.9 \\
& HAM-Net \cite{islam2021hybrid} (AAAI'21) & I3D & 65.4 & 59.0 & 50.3 & 41.1 & 31.0 & 20.7 & 11.1 & 49.4 & 30.8 & 39.8 \\
%& ACSNet \cite{liu2021acsnet} (AAAI'21) & I3D & - & - & 51.4 & 42.7 & 32.4 & 22.0 & 11.7 & - & 32.0 & - \\
%& LES \cite{liu2021weakly}, AAAI2021 & I3D & - & - & 50.8 & 41.7 & 29.6 & 20.1 & 10.7 & - & 30.6 & - \\
& WUM  \cite{lee2021weakly} (AAAI'21) & I3D & 67.5 & 61.2 & 52.3 & 43.4 & 33.7 & 22.9 & 12.1 & 51.6 & 32.9 & 41.9 \\
& AUMN \cite{luo2021action} (CVPR'21) & I3D & 66.2 & 61.9 & 54.9 & 44.4 & 33.3 & 20.5 & 9.0 & 52.1 & 32.4 & 41.5 \\
& CoLA \cite{zhang2021cola} (CVPR'21) & I3D & 66.2 & 59.5 & 51.5 & 41.9 & 32.2 & 22.0 & 13.1 & 50.3 & 32.1 & 40.9 \\
& TS-PCA \cite{liu2021blessings} (CVPR'21) & I3D & 67.6 & 61.1 & 53.4 & 43.4 & 34.3 & 24.7 & 13.7 & 52.0 & 33.9 & 42.6 \\
& UGCT \cite{yang2021uncertainty} (CVPR'21) & I3D & 69.2 & 62.9 & 55.5 & 46.5 & 35.9 & 23.8 & 11.4 & 54.0 & 34.6 & 43.6 \\
& ASL \cite{ma2021weakly} (CVPR'21) & I3D & 67.0 & - & 51.8 & - & 31.1 & - & 11.4 & - & - & - \\
% & CSCL \cite{ji2021weakly} (MM'21) & I3D & 68.0 & 61.8 & 52.7 & 43.3 & 33.4 & 21.8 & 12.3 & 51.8 & 32.7 & 41.9 \\
& CO$_2$-Net \cite{hong2021cross} (MM'21) & I3D & 70.1 & 63.6 & 54.5 & 45.7 & 38.3 & 26.4 & 13.4 & 54.4 & 35.6 & 44.6 \\
& D2-Net \cite{narayan2021d2} (ICCV'21) & I3D & 65.7 & 60.2 & 52.3 & 43.4 & 36.0 & - & - & 51.5 & - & - \\
& FAC-Net \cite{huang2021foreground} (ICCV'21) & I3D & 67.6 & 62.1 & 52.6 & 44.3 & 33.4 & 22.5 & 12.7 & 52.0 & 33.1 &  42.2 \\
& RSKP \cite{rskp} (CVPR'22) & I3D & 71.3 & 65.3 & 55.8 & 47.5 & 38.2 & 25.4 & 12.5 & 55.6 & 35.9 & 45.1 \\
& ASM-Loc \cite{he2022asm} (CVPR'22) & I3D & 71.2 & 65.5 & 57.1 & 46.8 & 36.6 & 25.2 & 13.4 & 55.4 & 35.8 & 45.1 \\
& Li et al. \cite{li2022forcing} (MM'22) & I3D & 69.7 & 64.5 & 58.1 & 49.9 & 39.6 & 27.3 & 14.2 & 56.3 & 37.8 & 46.1 \\
& DELU \cite{chen2022dual} (ECCV'22) & I3D & 71.5 & 66.2 & 56.5 & 47.7 & 40.5 & \textbf{27.4} & 15.3 & 56.5 & 37.4 & 46.4 \\
\cline{2-13}
& \textbf{Ours} & I3D & \textbf{74.0} & \textbf{69.4} & \textbf{60.7} & \textbf{51.8} & \textbf{42.7} & 26.2 & 13.1 & \textbf{59.7} & \textbf{38.9} & \textbf{48.3} \\
\hline
\hline
\end{tabular}}
\vspace{-4mm}
\end{center}
\end{table*}

\subsection{Comparison with State-of-the-art Methods}

In this section, we compare our method with state-of-the-art weakly supervised methods. Meanwhile, we also compare it several fully supervised methods. The results are shown in Table \ref{table:activity1.3} and Table \ref{table:THUMOS14}. On the THUMOS14 \cite{THUMOS14} benchmark, our method largely outperforms the previous weak-supervised approaches in almost every metric.
Besides, on the important criterions: average mAP (0.1:0.5), average mAP (0.3:0.7), average mAP (0.1:0.7), we surpass the state-of-the-art method, DELU \cite{chen2022dual}, by 3.2\%, 1.5\% and 1.9\%, respectively. Most excitingly, our method even outperforms some recent fully-supervised methods on the mAP (0.1:0.5) and average mAP (0.3:0.7) metrics. Some other weakly supervised methods (\emph{i.e.}, Weak \(\dagger\) in Table \ref{table:THUMOS14}) utilize additional weak supervisions, such as action frequency, our method still outperforms these methods. 

On the larger ActivityNet1.3 \cite{caba2015activitynet} dataset, our method outperforms all existing weakly supervised methods by a significant margin. Compared with previous methods, we consistently achieve a gain about 3.0\% on every metric. In terms of the average mAP, our method obtains a 3.7\% gain.

% To summarize, we conclude that our method is effective in the weakly supervised action localization task.

\subsection{Ablation Study}

Our ablation studies are conducted on the THUMOS14 benchmark. Unless explicitly stated, we follow our default setting in Sec. \ref{sec:implementation_datails}.
\begin{table}[!t]
\begin{center}
\caption{Localization results of using different distributions for action instance fusion during training and testing. The non-maximum suppression (NMS) is used as \emph{baseline} method.}
\vspace{-3mm}
\label{table:distribution_ablation}
\resizebox{1.0\columnwidth}{!}{
\begin{tabular}{l|ccc|ccc}
\hline
\hline
\multirow{3}{*}{Distribution} & \multicolumn{6}{c}{mAP @ IoU} \\
\cline{2-7}
& \multicolumn{3}{c|}{Training} & \multicolumn{3}{c}{Testing} \\
% \cline{2-7}
& 0.5 & 0.7 & AVG & 0.5 & 0.7 & AVG \\
\hline
\hline
Baseline &  36.2 & 12.2 & 44.9 & 36.6 & 11.1 & 44.9 \\
\hline
Uniform & 24.2 & 8.1 & 31.2 & 33.8 & 9.7 & 41.5 \\
Exponential & 36.2 & 12.3 & 45.3 & 41.1 & 12.4 & 47.1 \\
$t$-distribution & 36.7 & 12.7 & 45.6 & 41.2 & 11.5 & 47.1 \\
Gaussian & 36.6 & 12.7 & 45.6 & 41.2 & 11.6 & 47.1 \\
\hline
\hline
\end{tabular}}
\vspace{-4mm}
\end{center}
\end{table}

\vspace{-4mm}
\paragraph{Gaussian Weighted Instance Fusion}
As the initiative procedure in our pipeline, the Gaussian Weighted Instance Fusion module is of great importance. As pointed out previously, besides the Gaussian distribution, other distributions are also suitable for our fusion strategy.
% \textcolor{red}{However, we argue that the Gaussian weighted Instance Fusion module reduces information loss compared with the traditional Non-maximum suppression (NMS) approach, and other distributions.}

% To solidify this claim, we verify it from two aspects: (1) During the training stage, if we use different distributions, does Gaussian help to create informative pseudo-labels? (2) When trained with the same pseudo-label generation method, does Gaussian predicts the most accurate action instances?

In Table \ref{table:distribution_ablation}, we conduct several experiments to verify the effectiveness of our fusion strategy during the training stage. There are four different distributions: Gaussian distribution, Uniform distribution, Exponential distribution, and T-distribution to be considered. All these experiments are conducted with the same set of candidate action instances. Meanwhile, the LinPro Pseudo Label Generation module and $\Delta$ pseudo-label is turned off. We use the non-maximum suppression (NMS) approach as our baseline method. As we can see, our method achieves the best performance using the Gaussian distribution, the gain is about 0.7\% compared with the baseline method. This is because Gaussian distribution generally exists in a stochastic process. We also note that the uniform distribution deteriorates the performance because it significantly deviates from the ground truth distribution. Although the $t$-distribution exhibits a similar performance with Gaussian, we do not encourage using it. Since using $t$-distribution has no analytic form of fusing formulas, we have to solve a transcendental equation through Newton iteration.

Similarly, we conduct experiments in Table \ref{table:distribution_ablation} for evaluating the effectiveness of our fusion strategy during testing. With exactly the same distributions in the previous experiments, we choose NMS as our baseline. Specially, the baseline model is trained without the LinPro Pseudo Label Generation module and the $\Delta$ pseudo labels for a fair comparison. From Table \ref{table:distribution_ablation}, we can see that the Gaussian distribution still achieves the best performance and acquire a gain as high as 2.3\% at testing time. Intriguingly, we notice that the exponential distribution and $t$-distribution achieve the same performance. We argue that this is because those three distributions are very similar when the hyper-parameter of temperature is low during testing.

Although our method is almost hyper-parameter-free, we still suffer from the impact of the template distribution parameter temperature $T$. For this reason, we study the temperature's impact on our method. We conduct the following experiments  for two aspects. (1) We study the temperature's impact during the training stage, these experiments are carried out with the default setting. To ensure fair comparison, we do not use the fusion strategy during testing. (2) We study how the temperature influences the post-processing during testing. Also, we use the default settings with the training temperature set as 0.1.

\begin{figure}[!t]
  \centering
  \includegraphics[width=1\columnwidth]{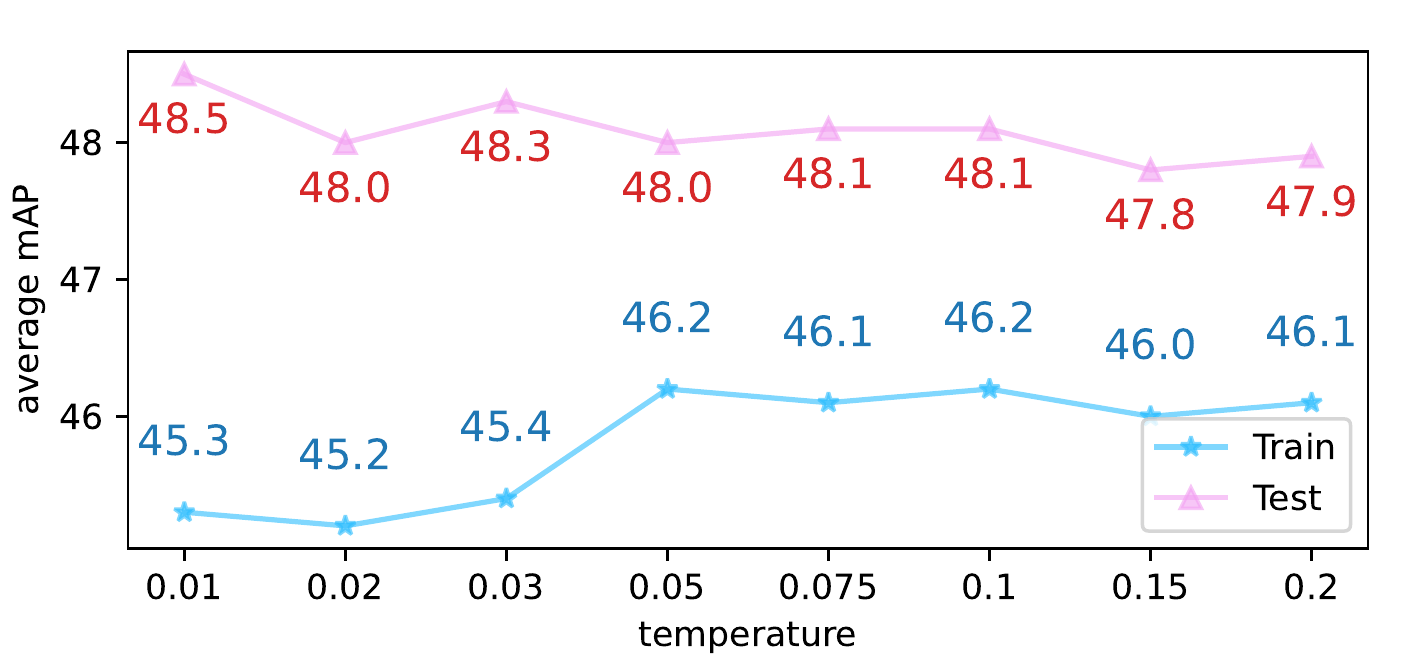}
  \caption{Impact of temperature on Gaussian weighted Instance Fusion module during training and testing.}\label{fig:temperature}
\vspace{-1mm}
\end{figure}

From the results in Figure \ref{fig:temperature}, we claim that our method is not sensitive to the hyper-parameter of temperature in a large range (from 0.05 to 0.2) during the training stage. Besides, our method suffers some performance loss when the temperature is too low. Under this circumstance, the candidate action instances of high sampling probability would be overwhelming during the fusion process, we cannot collect information from regions where the probability is low.

Similarly, during testing, {the temperature impact on our method is also limited. There is a large region where our method doesn't change much as the temperature varies. One may notice that our method achieved 48.5\% in terms of mAP as temperature 0.01, however, we do not report it as our main result because we believe this temperature is too radical. Instead, we choose temperature 0.03 as our default setting since it lies in the center of our temperature interval}.

\begin{table}[!t]
\begin{center}
\caption{Evaluation of each component of our method.}
\vspace{-3mm}
\label{table:componet_ablt}
\resizebox{1.0\columnwidth}{!}{
\begin{tabular}{l|ccccc}
\hline
\hline
\multirow{2}{*}{componet} & \multicolumn{4}{c}{mAP @ IoU} \\
\cline{2-5}
& 0.3 & 0.5 & 0.7 & AVG \\
\hline
\hline
baseline & 55.8 &38.2 & 12.5 & 45.1 \\
\hline
+ LinPro Pseudo Label Generation & 57.4 & 36.2 & 12.2 & 44.9 \\
+ Gaussian weighted fusion & 58.2 & 36.6 & 12.7 & 45.6 \\
+ $\Delta$ pseudo label & 58.4 & 37.6 & 12.2 & 46.2 \\
\hline
+ Gaussian weighted fusion (testing)& 60.7 & 42.7 & 13.1 & 48.3 \\
\hline
\hline
\end{tabular}}
\vspace{-7mm}
\end{center}
\end{table}
% \begin{figure}[!t]
%   \centering
%   \includegraphics[width=0.9\columnwidth]{em_iter_num.pdf}
%   \caption{The detection results of two implementations of Equations (\ref{eq:birw1}) and (\ref{eq:birw2}) with different iteration numbers.}\label{fig:em_iter_num}
% \vspace{-2mm}
% \end{figure}

\vspace{-4mm}
\paragraph{Performance gain of each component.}
Our method consists of three components, \ie, Gaussian weighted instance fusion, LinPro pseudo label generation, and $\Delta$ pseudo label. Here, we study the performance gain of each component and their combinations, we take the RSKP \cite{rskp} as our baseline method, and gradually add the three components. As we can see in Table \ref{table:componet_ablt}, when only adopting the LinPro pseudo label generation to the NMS-generated action boundaries, the performance drops, indicating the NMS cannot generate high-quality action boundaries. After adding the Gaussian weighted fusion module, the performance increases by 1.5\% over the baseline. Further benefitted from the introduction of $\Delta$ pseudo labels, our method can reach the average mAP of 46.2\%. 
Finally, an evident gain of 2.1\% can be obtained when the Gaussian weighted fusion strategy is adopted as post-processing during testing.

\begin{table}[!t]
\begin{center}
\caption{The detection results of applying our method to existing methods. \emph{Embedding} means we add a learnable network after the backbone network to learn the video features. The results of the original methods are reproduced.}
\vspace{-3mm}
\label{table:improve_existing_methods}
\resizebox{1\columnwidth}{!}{
\begin{tabular}{l|ccccl}
\hline
\hline
\multirow{2}{*}{Method} & \multicolumn{4}{c}{mAP @ IoU} \\
\cline{2-5}
& 0.3 & 0.5 & 0.7 & AVG \\
\hline
\hline
STPN \cite{nguyen2018weakly} + embedding & 38.4 & 19.1 & 4.7 & 28.4 \\
STPN + embedding + Ours & 41.6 & 22.1 & 6.8 &31.6 $_{\uparrow 2.2}$ \\
\hline
BM \cite{nguyen2019weakly} & 45.2 & 26.2 & 8.7 & 35.3 \\
BM + Ours & 47.4 & 28.5 & 10.3 & 37.3 $_{\uparrow 2.0}$\\
\hline
WUM \cite{lee2021weakly} & 51.0 & 32.8 & 10.9 & 40.4 \\
WUM + Ours & 53.2 & 34.0 & 11.4 & 42.3 $_{\uparrow 1.9}$\\
\hline
FAC-Net \cite{huang2021foreground} & 53.2 & 34.4 & 13.7 & 42.9 \\
FAC-Net + Ours & 56.4 & 37.2 & 13.0. & 44.8  $_{\uparrow 1.9}$ \\
\hline
\hline
\end{tabular}}
\vspace{-2mm}
\end{center}
\end{table}

\vspace{-4mm}
\paragraph{Integrating our modules to existing methods.}
In Table \ref{table:improve_existing_methods}, we plug our proposed pipeline into some existing methods. The default settings of these methods are used for fair comparisons. As we can see, our method can consistently improve the performances of the previous methods.
\begin{figure}[!t]
  \centering
  \includegraphics[width=0.88\columnwidth]{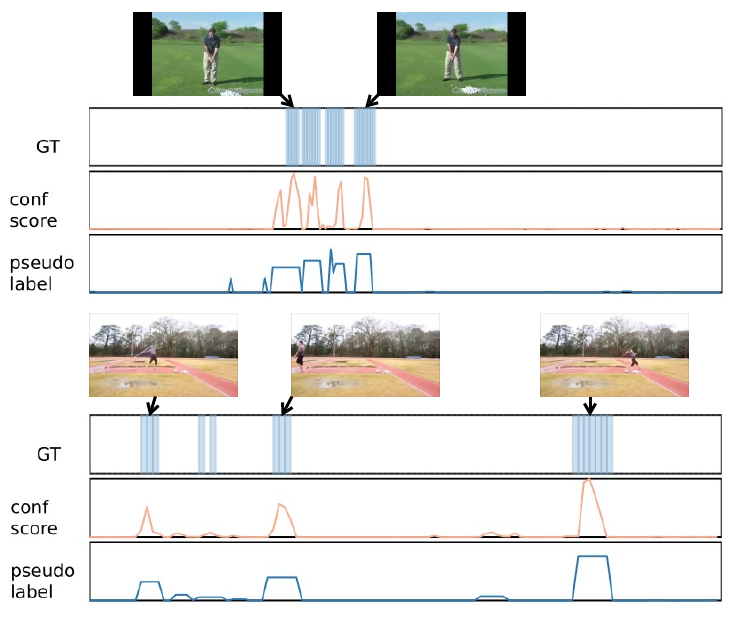} 
  \caption{Illustration of the detection results of \emph{GolfSwing} and \emph{JavelinThrow}. GT stands for the ground truth, conf score is the predicted confidence score of the given action. The pseudo labels are generated by our Linpro pseudo label generation module.}   \label{fig:visualization}
\vspace{-3mm}
\end{figure}

\paragraph{Qualitative results.}
We visualize some examples of the confidence scores and Linpro pseudo labels generated by our method. As we can see from Figure \ref{fig:visualization}, the generated pseudo labels preserve the confidence score within each action instances, meanwhile they are as uniform as possible. 

\section{Conclusion}
\vspace{-1mm}
In this paper, we propose a novel framework to generate better pseudo labels from action boundaries.
We first propose a Gaussian weighted instance fusion module to obtain high-quality action boundaries. After that, we generate the pseudo labels from action boundaries by solving the the optimization problem under the constraints in terms of the confidence scores of action instances.
Finally, we propose to utilize the $\Delta$ pseudo-label for introducing a self-correction mechanism into the model. Our method achieves state-of-the-art performance on THUMOS14 and ActivityNet1.3 and can consistently improve the performance of existing methods.

\section{Acknowledgments}
This project is funded in part by National Key R\&D Program of China Project 2022ZD0161100,  by the Centre for Perceptual and InteractiveIntelligence (CPII) Ltd under the Innovation and TechnologyCommission (ITC)'s InnoHK, by General Research Fund Project 14204021 and Reseasrch Impact Fund Project R5001-18 of Hong Kong RGC.
%%%%%%%%% REFERENCES
{\small
\bibliographystyle{ieee_fullname}
\bibliography{egbib}
}

\end{document}